%% file: main.tex
\documentclass[10pt, conference]{IEEEtran}
\usepackage[utf8]{inputenc} 
\usepackage{amsmath,amssymb,amsfonts}
\usepackage{graphicx}
\usepackage{textcomp}

\usepackage{amsthm}
\usepackage{xcolor}
\usepackage[figuresright]{rotating}

\usepackage{graphicx}
\graphicspath{{/}{fig/}}

\usepackage{array}
\usepackage{textcomp}
\usepackage{xcolor}
\usepackage{multirow}
\usepackage{booktabs}

\usepackage{mathtools}
\usepackage{breqn}
\usepackage{float}

\usepackage{pgfplots}
\pgfplotsset{compat=1.7}
\usepgfplotslibrary{groupplots}

\usepackage{multirow,tabularx}
\usepackage{flushend}
\usepackage{hyperref}

\usepackage[style=base]{caption}
\usepackage{subcaption}

\newcommand{\red}{\textcolor{red}}

\newcommand{\linh}[1]{\textbf{\color{red} Linh: #1}}

\newlength\figureheight
\newlength\figurewidth
\title{\LARGE \bf
    End-to-End Design for Self-Reconfigurable \\ Heterogeneous Robotic Swarms
}

\author{
    \IEEEauthorblockN{
        Jorge Peña Queralta\textsuperscript{1},
        Li Qingqing\textsuperscript{1},
        Tuan Nguyen Gia\textsuperscript{1},
        Hong-Linh Truong\textsuperscript{2},
        Tomi Westerlund\textsuperscript{1}
    }
    \IEEEauthorblockA{
        \textsuperscript{1} \href{https://tiers.utu.fi}{Turku Intelligent Embedded and Robotic Systems (TIERS) Lab, University of Turku, Turku, Finland} \\
        \textsuperscript{2} Department of Computer Science, Aalto University, Finland \\
        Emails: \textsuperscript{1}\{jopequ, qingqli, tunggi, tovewe\}@utu.fi, \textsuperscript{2} linh.truong@aalto.fi \\
    }
}

\begin{document}

\maketitle
\thispagestyle{empty}
\pagestyle{empty}

\global\csname @topnum\endcsname 0
\global\csname @botnum\endcsname 0
\input{sections/00_Abstract.tex}
\input{sections/01_Introduction.tex}

\input{sections/02_Reconfigurability.tex}
\input{sections/03_Applications.tex}

\input{sections/04_Conclusion.tex}


\bibliographystyle{IEEEtran}
\bibliography{main.bib}

\end{document}

%% file: sections/00_Abstract.tex
\begin{abstract}

    More widespread adoption requires swarms of robots to be more flexible for real-world applications. Multiple challenges remain in complex scenarios where a large amount of data needs to be processed in real-time and high degrees of situational awareness are required. The options in this direction are limited in existing robotic swarms, mostly homogeneous robots with limited operational and reconfiguration flexibility. We address this by bringing elastic computing techniques and dynamic resource management from the edge-cloud computing domain to the swarm robotics domain. This enables the dynamic provisioning of collective capabilities in the swarm for different applications. Therefore, we transform a swarm into a distributed sensing and computing platform capable of complex data processing tasks, which can then be offered as a service. In particular, we discuss how this can be applied to adaptive resource management in a heterogeneous swarm of drones, and how we are implementing the dynamic deployment of distributed data processing algorithms. With an elastic drone swarm built on reconfigurable hardware and containerized services, it will be possible to raise the self-awareness, degree of intelligence, and level of autonomy of heterogeneous swarms of robots. We describe novel directions for collaborative perception, and new ways of interacting with a robotic swarm.

\end{abstract}


\IEEEpeerreviewmaketitle

%% file: sections/01_Introduction.tex
\section{Introduction}

\begin{figure}
    \centering
    \includegraphics[width=0.42\textwidth]{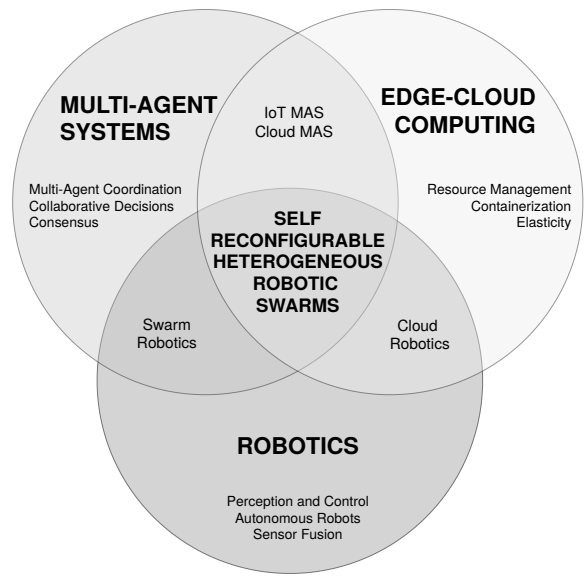}%
    \caption{Intelligent and self-reconfigurable robot swarms can be designed at the intersection of the multi-agent systems (MAS), Robotics and Edge-Cloud Computing domains.}
    \vspace{-0.8ex}
    \label{fig:gap}
\end{figure}

A recent trend in cyber-physical systems and the Internet of Things (IoT) domain is to shift towards more distributed computation, a trend that has crystallized through the edge computing paradigm~\cite{8789742}. Similarly, recent advances in containerization, elastic computing, and dynamic resource management are materializing a decentralized cloud~\cite{Ferrer:2019:TDC:3303862.3243929, Jalali:2017:CIG:3123878.3132008}. Multiple researchers have explored the possibilities of integrating Multi-Agent Systems (MAS) theory within the IoT and cloud computing towards IoT MAS and Cloud MAS~\cite{calvaresi2017challenge}. On the other side, the combination of MAS and robotics has brought swarms of robots with the potential for enhancing human responses in safety-critical applications such as firefighting~\cite{penders2011robot}, or post-disaster scenarios~\cite{de2017incremental}, among others.

We extend these two approaches (swarm robotics and IoT MAS) towards the intersection of edge computing and robotics. and argue that with appropriate management and distribution of computing, sensing and communication resources within a robotic swarm, higher degrees of intelligence and operational flexibility and robustness can be achieved. This concept is illustrated in Fig.~\ref{fig:gap}. The capabilities of robotic swarms are currently limited by different factors, from the lack of methods and distributed collaborative sensing algorithms~\cite{masehian2017cooperative}, to static and inflexible resource management with non-optimal utilization of hardware resources due to separate hardware and software design. This includes embedded hardware with relatively constrained computational resources due to payload constraints and uniform resources in swarms~\cite{kattepur2016resource, hong2019qos}. Some previous works have addressed these limitations through computational offloading and the definition of edge-cloud robotics architectures~\cite{qingqing2019odometry}. More recently, heterogeneous robotic swarms have been proposed to extend the flexibility and intelligence, opening the door to a wider array of more complex application possibilities~\cite{masehian2017cooperative}. Nonetheless, multiple challenges remain in terms of managing the collaboration within heterogeneous swarms~\cite{queralta2020blockchain}. 

\subsection{Related Works}

Multiple research efforts within the fields of multi-robot systems and cloud computing have been directed towards distributed task allocation and distributed load management. For instance, autonomous mobile programs (AMPs) were an early introduction of a dynamic computational load management framework~\cite{deng2006autonomous}. AMPs provided a distributed approach where autonomous agents were able to make decisions on a shared computational load, being aware of their own computational capabilities. AMPs share similarities with early load balancing techniques based and colony optimization for cloud computing~\cite{zhang2010load}. More recently, multi-agent load balancing for resource allocation in a distributed computing environment has been proposed~\cite{banerjee2017multi}. From the point of view of task allocation in multi-robot systems, K-means clustering and auction based mechanisms were introduced in~\cite{elango2011balancing}. In terms of spatial allocation, a workspace partitioning method was presented in~\cite{gautam2017balanced} for indoor environments. In our work, we aim at combining these two approaches considering full reconfigurability through tight integration of methods form the edge computing domain and algorithms for cooperation in multi-robot systems. This is, to the best of our knowledge, the first paper presenting such approach.

In our recent works, we have presented initial results towards the definition of reconfigurable robotic swarms. In~\cite{asrepreprint}, we define the concept of resource ensembles from the perspective of the edge computing domain. This concept serves as the basis for abstracting edge resources and building dynamic management models on top of them. From the point of view of collaborative swarms of robots, we have presented a blockchain-based approach in~\cite{queralta2020blockchain}. In this work, we utilize a blockchain as a medium for achieving consensus for bandwidth allocation and data quality ranking in a distributed multi-robot system. This, again, serves as the starting point towards distributed sensing and data processing in swarms of robots, where sensing, network and computational resources are abstracted and managed through collaborative decision making.

\subsection{Contribution and Structure}

The techniques we propose have a clear impact on swarms of drones. Current solutions for drone swarms require operators to either manually control drones or perform analysis of streamed data at a ground control center. This is a limitation for the deployment of drones in large areas where there have been natural disasters such as fires, or where people have gone missing, as the human resources necessary are too large. 
Even if the data is processed by a computer at the ground station, the need for a high-bandwidth channel between drones and the base station still limits significantly their operational capabilities. Therefore, there is an evident need for more intelligent drones that are able to perform data analysis independently and autonomously navigate large areas. 
Reconfigurable drone swarms empower complex edge data analysis at the swarm level through distributed edge computing. At the same time, this allows for long-term autonomous operation when energy constraints allow, as well as reconnaissance in remote areas with poor network connectivity.

The main objective of this paper is to introduce a {\em new design approach} that enables efficient and dynamic resource management and autonomous reconfiguration of heterogeneous robotic swarms (Sections \ref{sec:models} \& \ref{sec:architecture}). We discuss the optimization of the various computing resources and sensing capabilities of robots in the swarm through a hardware (HW)/software (SW) co-design approach for the development of specialized robots. In particular, we incorporate elastic principles for coordinating and engineering collective capabilities of multiple heterogeneous robots. This elasticity takes into account the coordination at the level of multi-agent systems, but also the specific resources and algorithms utilized for robotic perception and navigation, among others.

Furthermore, we discuss how our proposed approach enables the reconfiguration of drone swarms and realize a distributed collective intelligence (Section \ref{sec:app}). This requires embedding intelligence and information processing in the drones themselves. With current technology, deploying multiple drones requires coordination among their operators, binding valuable resources from the actual mission. Even more, when multiple drones are being deployed in parallel~\cite{gregory2016application, hu2018joint}. However, neither isolated intelligent drones nor simple task list orchestration is sufficient~\cite{mendes2018flexible, lomonaco2018intelligent}.  Thus, it is essential to establish a collective intelligence that enables autonomous coordination and collaboration among the drones.

The remainder of this paper is organized as follows. Section II describes the main concepts involved in swarm reconfigurability, interfacing and control. In Section III, we introduce a three-layered end-to-end design architecture for reconfigurable robotic swarms. Section IV then delves into enabled technologies, focusing on our initial work towards a reconfigurable drone swarm. In Section V, we discuss potential application scenarios. Section VI concludes the work.


%% file: sections/02_Reconfigurability.tex
\begin{figure}
    \centering
    \begin{subfigure}[t]{0.48\textwidth}
        \centering
        \includegraphics[width=\textwidth]{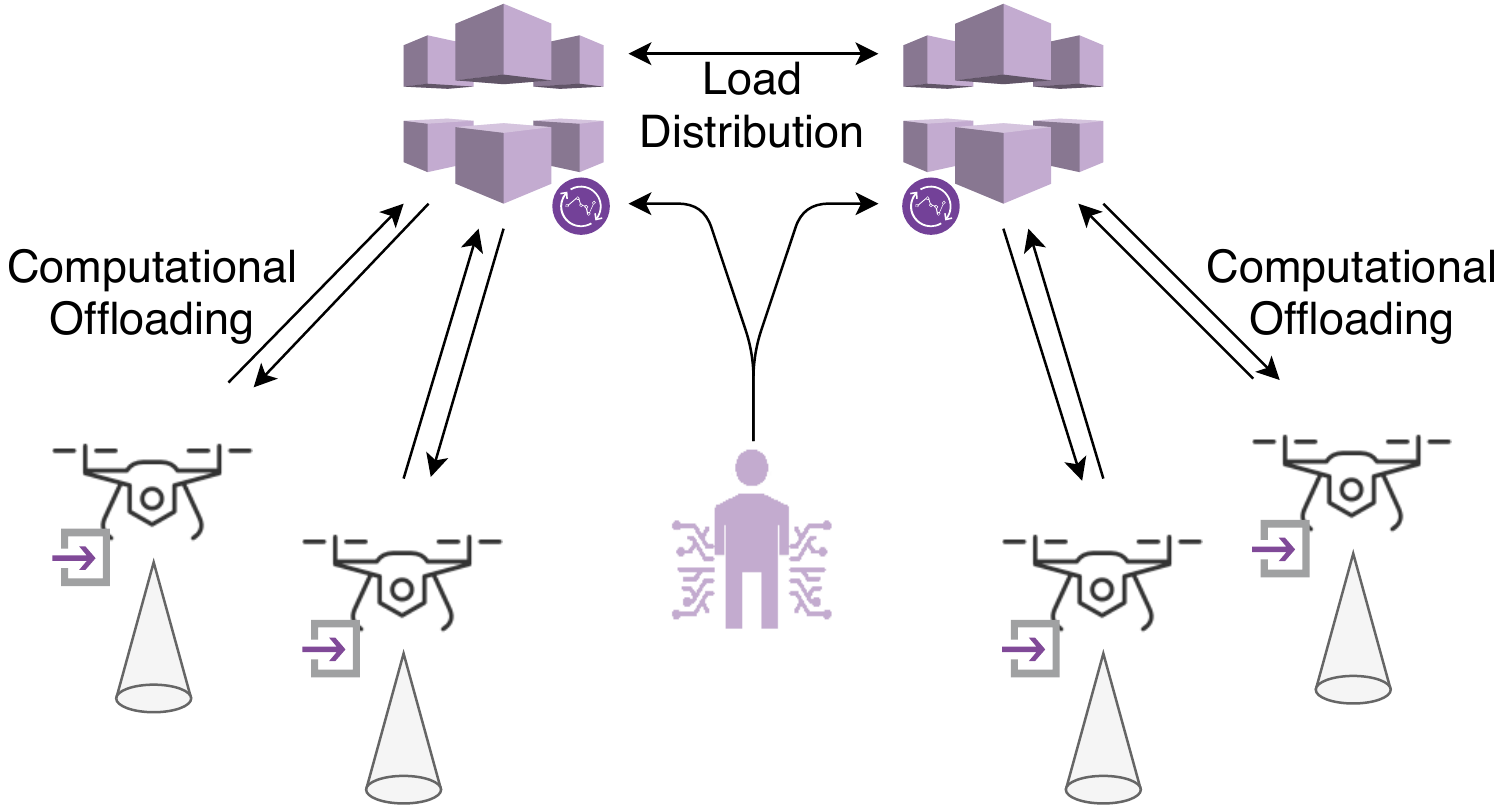}
        \caption{A cloud robotics system.}
        \label{subfig:scenario1}
    \end{subfigure}
    \vspace{1ex}
    \begin{subfigure}[t]{0.48\textwidth}
        \centering
        \includegraphics[width=\textwidth]{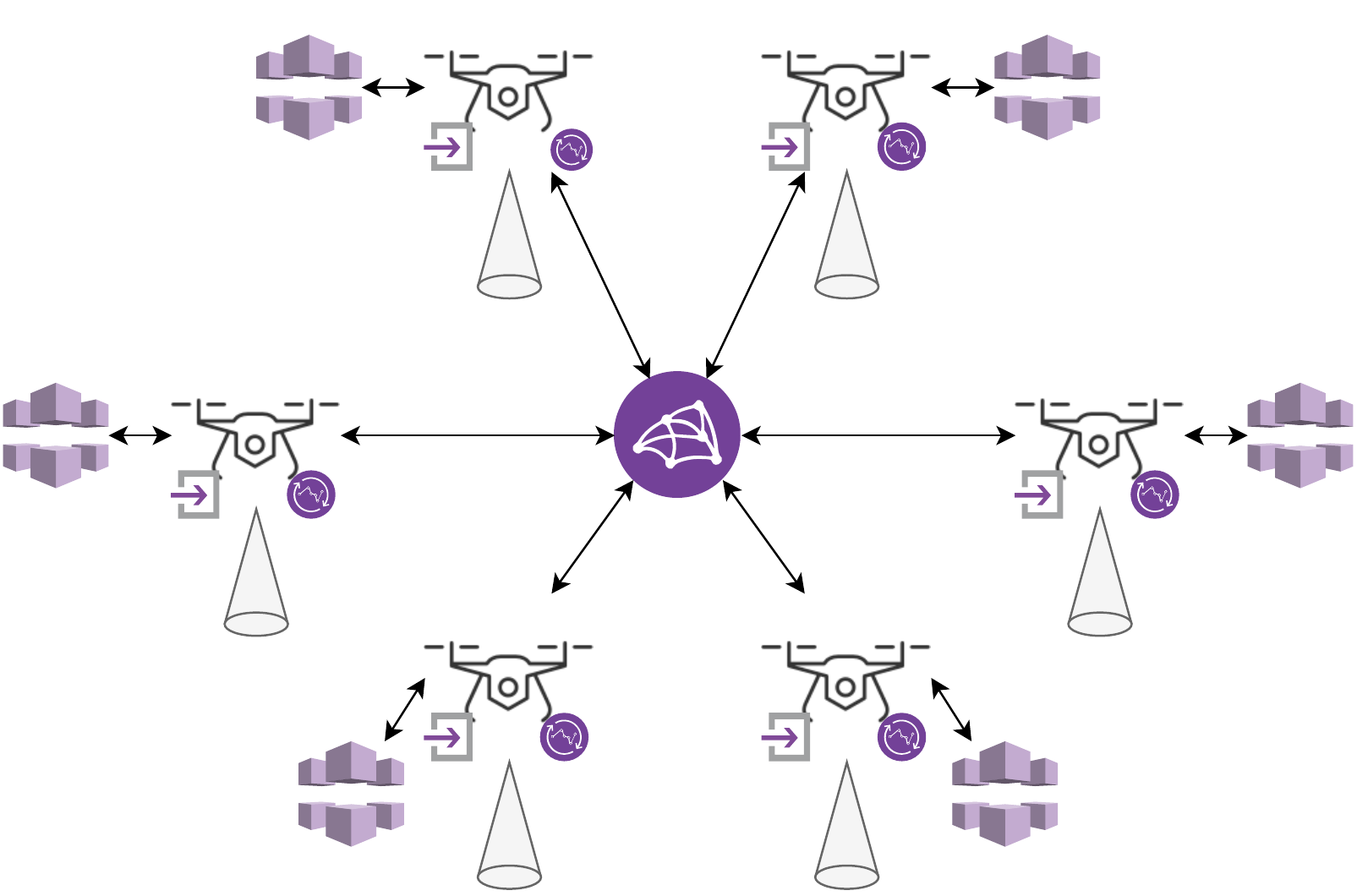}
        \caption{A swarm robotics system.}
        \label{subfig:scenario2}
    \end{subfigure}
    \vspace{1ex}
    \begin{subfigure}[t]{0.48\textwidth}
        \centering
        \includegraphics[width=\textwidth]{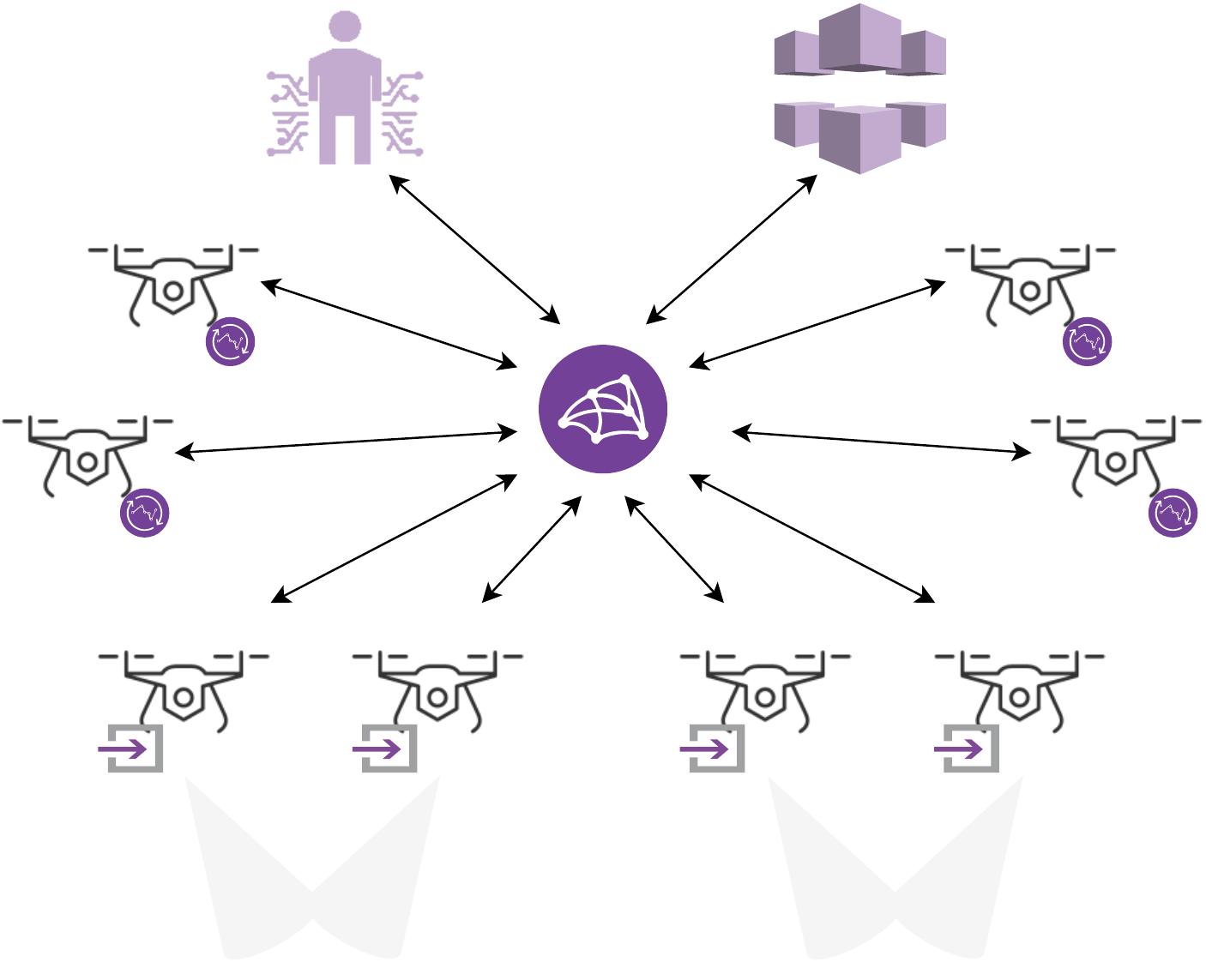}
        \caption{A reconfigurable robotic swarm.}
        \label{subfig:scenario3}
    \end{subfigure}
    \vspace{1ex}
    \includegraphics[width=0.42\textwidth]{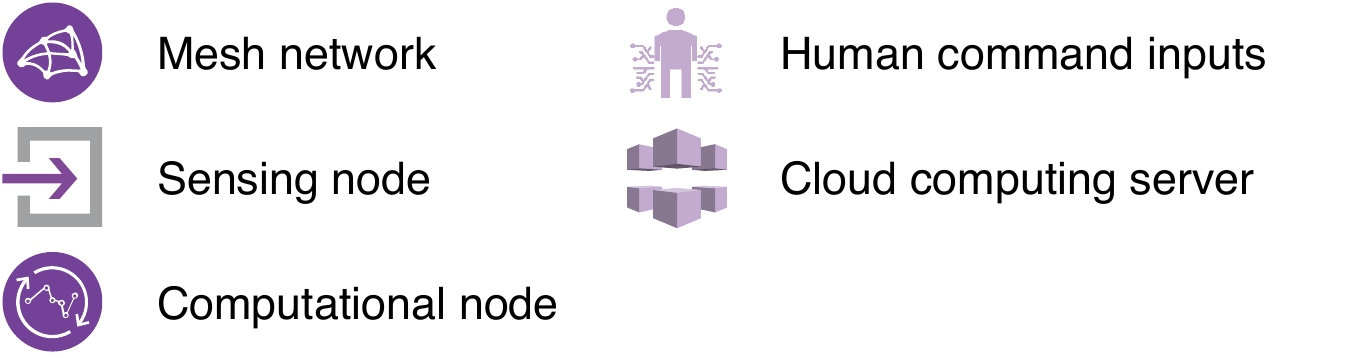}
    \caption{Illustration of different connectivity modalities for a swarm of drones, illustrating sensing and computing roles together with interaction modalities.}
    \label{fig:scenarios}
\end{figure}

\section{Models}
\label{sec:models}

We address the aforementioned challenges through an end-to-end design: from the robot hardware and local decision making on the computational and sensing resources to providing the swarm and its capabilities as a service for end-users. Two key aspects in our work are:
\begin{itemize}
    \item Reconfigurable hardware resources for flexible and resource-rich computation platform at the swarm level
    \item Distributed data processing at the edge for collective swarm intelligence.
\end{itemize}
Reconfigurability and distributed intelligence provides more computational resources for multi-modal sensor fusion and distributed computation. Our model leverages methodologies and engineering techniques for distributed and dynamic management of elastic resources for swarms improving quality of results. The models are:

\noindent {\bf 1) Swarm-as-a-Service Model:}  A swarm provides services to end-users with a control interface or API through what we call a Swarm as a Service (SwaaS). SwaaS offers an edge service model for swarm applications. In this view, each robot with its specialization is an edge services provider (or just an edge provider). Edge provider's sensing, computational and external communication resources are considered edge resources. Edge resources (hardware and software) are co-designed to make robots richer in terms of the resources that they can provide. With this approach, we bring various concepts of edge computing and services models to the field of multi-robot systems.

\noindent  {\bf 2) Application-Specific Resource Ensembles Model}: Application Specific Resource Ensembles (ASREs) \cite{asrepreprint} define a specific organized set of edge resources (ASRE template or pattern) in swarms forming the edge infrastructure. An essential part of ASREs is the coordination and monitoring of resources together with swarm control and coordination. Resource management techniques enable service discovery, service end-to-end communication segmentation, and distributed task computation under unreliable and uncertain environments by incorporating uncertainty and elasticity~\cite{truong2017modeling}. For our knowledge, these have not beed applied before to drone swarms. For example, \cite{7881642, 216781} are dedicated for containers and virtual machines. 

In our end-to-end vision, the central point is to provide dynamic and flexible swarms as an \textit{elastic heterogeneous multi-robot system}. In the system individual robots have different sensing and computational capabilities, a mesh network takes care of intra-swarm communication, and distributed algorithms enable the swarm to perform collective decision making as if it were a single unit. By \textit{elastic}, we mean that the different resources of robots are abstracted and can be reconfigured depending on the application needs~\cite{DBLP:journals/internet/DustdarGST11}. An illustration of this concept, compared to current cloud robotic and swarm robotic systems, appears in Fig.~\ref{fig:scenarios}. In Fig.~\ref{subfig:scenario1}, each drone is independently connected to a cloud server, where it offloads part of its data processing. Any external control in this case goes through the cloud application, but direct control of drones could be enabled as well (for example, if the movement of the drone is controlled via a radio controller, and then mapping or other algorithms are run in the cloud). In Fig.~\ref{subfig:scenario2}, the drones form together a swarm. Each individual drone in the swarm has the same role initially, and both sensing and data processing occur individually at each drone. Algorithms describing the collaboration between drones could then run at the swarm level, but each drone would be still a separate entity. Finally, our approach is illustrated in Fig.~\ref{subfig:scenario3}, where all drones form a swarm as well. The key difference is that the swarm and its applications and resources are abstracted from individual drones and defined in a distributed manner at the swarm level. The communication with a controller or cloud services occurs from the swarm as a whole, and not from individual drones as separate entities. Moreover, sensing and computational resources, and the corresponding roles, are assigned dynamically among the swarm members. In the example illustrated in the figure, half of the drones take a sensing role as their main role while data processing is offloaded to other drones assigned as computing nodes.

\section{Architecture Layers}
\label{sec:architecture}

\begin{figure*}[t]
    \centering
    \includegraphics[width=0.99\textwidth]{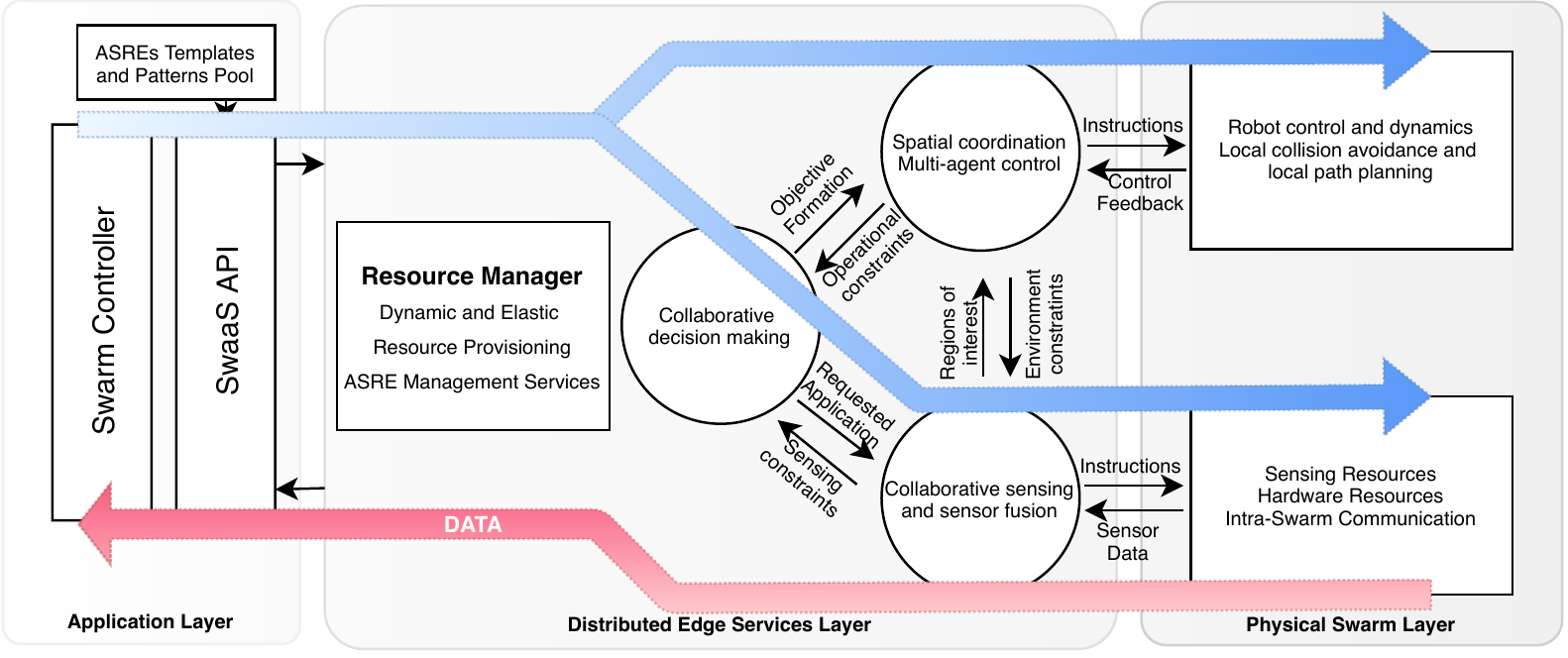}
    \caption{Proposed Architecture and Building Blocks for SwaaS and ASREs}
    \label{fig:arch}
\end{figure*}

We design an architecture with  
three layers and building blocks as illustrated in Fig.~\ref{fig:arch}. The layers are: 

{\bf Physical Swarm Layer.} The actual control of robots is carried out at the physical layer, where the different hardware and mesh communication solutions are defined. The definition of a set of computing resources, sensors and actuators needs to be carried out as part of the swarm design in order to enable higher degrees of optimization when the sensing and computational resources are shared at the swarm level and applications run in a distributed manner.

{\bf Edge Services Layer}. This is the main focus of the swarm design in terms of self-reconfigurability. This layer includes a Resource Manager (RM) for configuring and managing resources provided by the Physical Swarm Layer. Based on application requirements, the RM will assemble resources from edge providers into ASREs for SwaaS. The RM supports the automatic creation of ASREs by requesting, provisioning and orchestrating suitable resources. 

This layer represents all the distributed services and processes running within the swarm and executed at each individual robot. These processes are classified in three main types: (1) spatial coordination (e.g., distributed formation control~\cite{queralta2019indexfree, mccord2019progressive}), (2) collaborative sensing (e.g. cooperative mapping~\cite{queralta2019collaborative}), and (3) collaborative decision making (e.g. ). These three apparently different parts of swarm control and decision making have a high synergy and their optimal operation depends on feedback from each other. These three topics have mostly been studied separately in the previous works~\cite{oh2015survey, ccms, mavrovouniotis2017survey}. Therefore, we have focused on the design and development of techniques for efficient communication between these processes. In summary, the key novelty is that ASREs and the RM implicitly manage collaboration within the swarm.

At runtime, ASREs will form an elastic and resilient edge mesh of services across robots in a swarm. The RM will dynamically provision new resources from different providers elastically. This kind of elasticity is carried out in an end-to-end and bi-directional manner: the resources are provisioned dynamically when new services are required or when the available resources change. The RM learns and optimizes the provisioning based on the reliability of resources, performance variations, bottlenecks, and failures. The RM provisioning is hidden from the application. 

{\bf Application Layer.} The application layer provides an interface for controlling and interacting with the swarm. We refer to this interface as the Swaas API. An external party or swarm controller can select from a pool of ASRE templates, which define the different patterns in which the swarm can be configured for different applications. By choosing an ASRE template through the SwaaS API, the swarm controller is implicitly selecting a set of resources and services. These resources are then provisioned and managed within the swarm itself through the RM. The services are provided based on the available distributed algorithms for sensing and coordination.



%% file: sections/03_Applications.tex
\section{Reconfiguration Processes in a Drone Swarm}
\label{sec:app}

In this section, we discuss the specific technologies that enable the realization of the self-reconfigurable robotic swarm architecture defined in the previous section. We are in the process of applying these technologies to a swarm of drones.

\subsection{Reconfigurability-enabling Technologies}

Currently, most small mobile robots and aerial drones rely mainly on CPUs in order to perform all the navigation and mission related computation, and microcontrollers for low-level control such as flight controllers~\cite{kaufmann2019beauty}. We are leveraging existing technologies from other domains to enable reconfiguration within a drone swarm.

As a computing platform, we utilize FPGAs with embedded processors to extend the existing algorithms with custom hardware accelerators. FPGAs are reconfigurable hardware accelerators that can be exploited in computationally intensive and highly parallelizable tasks for autonomous robots, with higher performance/size and performance/power ratio as we have shown in previous works~\cite{qingqing2019fpga, queralta2019sensors}. Both the size and power consumption of hardware are essential aspects to take into account in drones. The use of FPGAs enables the RM to not only provision the existing resources but also dynamically provision new hardware accelerators on-demand. Some drones are equipped with FPGAs while others have embedded processors with GPU such as the NVIDIA Jetson TX2. 

At the software level, we utilize containers to enable dynamic resource management and task execution. Container technologies are known but they have not been exploited for drones. Our goal is to use containers to enable dynamic resources management and task execution.  All algorithms, from spatial coordination to collaborative sensing, are containerized and run in a distributed way. With efficient container orchestration, we are able to add flexibility and reconfigurability to the swarm. We bring specific techniques for computational load distribution, elasticity and resource management from the edge-cloud domain to the robotics domain. 

In order to interface sensors, actuators, communication and the containerized algorithms, we utilize Robot Operating System (ROS2) which runs as a container application as well. ROS is the de-facto standard for robotic development in both academia and industry. ROS2 focuses on distributed multi-robot systems and real-time computing, and will allow us to exploit container technologies for drones.

For network interfacing, there are no solutions that integrate ROS2 and mesh networking so far. Our experiments will utilize more traditional solutions at first, with all drones connected to a single Wi-Fi access point. Nonetheless, we will work towards the integration of ROS2 and a Bluetooth 5 mesh network. Another recent technology that can provide significant advantages is ultra-wideband (UWB)~\cite{shule2020uwbbased}. UWB enables accurate localization in multi-robot systems and, in particular, in drones~\cite{queralta2020uwb}, and has the potential for simultaneous communication and localization. We will work on extending our current works on UWB-based mobile localization systems~\cite{almansa2020autocalibration}, studying the integration of UWB as a network interface between ROS2 nodes.

Finally, we leverage blockchain-powered consensus algorithms suitable for multi-robot systems. Recent works \cite{strobel2018managing, queralta2020blockchain, ferrer2018blockchain} have proposed design concepts for integrating next-generation low-latency and scalable blockchains within heterogeneous multi-robot systems. Blockchains can be utilized as a distributed framework to achieve consensus in a multi-robot system, and also validate identities. This can be then utilized by ASRE management services, which could, in turn, be implemented as distributed Smart Contracts for resource coordination. The idea of utilizing a blockchain-based framework for managing edge resources has already been explored in our previous works~\cite{queralta2020enhancing}. 



\subsection{Edge computing algorithms for sensing}

We classify the algorithms in the edge layer in three main types: spatial coordination, collaborative sensing, and decision making. In previous works, these are typically defined with strong dependencies. For instance, depending on the sensing variable robots might be required to be in a specific spatial formation~\cite{queralta2019progressive}. However, in our architecture these algorithms are designed independently and abstracted as edge services. This modular architecture brings multiple advantages. For instance, spatial coordination algorithms take feedback from the sensing algorithms regarding the location of regions of interest that should be analyzed more closely. This feedback is used to rearrange the drones in the proper shape and location. At the same time, the role of each drone within the spatial pattern is given by the collaborative decision making process. Analogously, the spatial coordination algorithms give feedback to other processes about the movement constrains of the swarm surroundings. 


\subsection{Resource Management}

The RM provisions and manages all resources, from hardware to edge services. Sensing resources are abstracted through ROS nodes (drivers) that produce data in standard formats. Each edge service is broken down into sub-services (for example, independent parts of an algorithm) and each of these is abstracted as a ROS node that consumes and produces different types of data (always in standard formats). All these ROS nodes are containerized and provisioned by the RM across the available the computing resources. The computing resources are modeled based on the amount and type of containers that they can run, and the performance for each containers. The provisioning is an optimization process that takes into account communication latency between data producers and consumers, and execution latency. The RM itself runs as a distributed and containerized application across the swarm, and manages the resources with elastic techniques.

Each application that utilizes the SwaaS API must define a Quality of Results (QoRs) requirement. We bring this concept from elastic computing models in which a QoR is defined in terms of performance, quality of data, type of output, and other measurable information~ \cite{Truong:2014:PSE:2624303.2624633}, which the swarm provides to the controller through the SwaaS API. The QoR is essential for elastic resource management to dynamically provision the different resources taking the QoR requirement as an optimization constraint. We will extend the work by Mariani \textit{et al.} on coordination-aware elasticity for developing primitives and algorithms to control the elasticity of swarms~\cite{Mariani:2014:CE:2759989.2760093}.

\section{Application Scenarios}

Drones are already being utilized by fire departments~\cite{price2019public}, police forces~\cite{miller2016use}, and for creating ad-hoc networks in post-disaster scenarios~\cite{panda2019design}. Fire and police departments are using drones for monitoring and surveying emergency scenes and disaster sites as well as finding missing people. Going beyond perception-centric usage would be beneficial in all of these use cases. This would require embedding intelligence and information processing in the drones themselves, especially when authorities are starting to deploy multiple drones in parallel for the same applications. Deploying multiple drones requires coordination and collaboration among their operators binding valuable resources from the actual mission. Therefore, it is essential to establish a collective intelligence that enables autonomous coordination and collaboration among the drones instead of passively receiving the tasks individually from a centralized control system. 


Another key application area for UAVs is real-time surveying and monitoring. The same methods and technologies presented in this paper can be applied in different tasks such as environmental monitoring, land surveying or scientific study of different parameters in remote locations. The main benefit of utilizing reconfigurable drone swarms is the ability to perform complex data analysis at the Swarm level through distributed Edge computing. This allows for long-term autonomous operation when energy constraints allow, as well as reconnaissance in remote areas with poor network connectivity.




%% file: sections/04_Conclusion.tex
\section{Conclusion}

We have proposed an architectural definition for reconfigurability in heterogeneous robotic swarms. This architecture is based on a combination of concepts and techniques from the robotics domain, multi-agent systems domain and edge-cloud computing domain. This is, to the best of our knowledge, the first work that proposes the abstraction and management of both hardware (sensors, actuators, computation and communication) and software (distributed sensing, coordination and decision making) as edge resources with elastic techniques. In particular, we explain how we are designing a reconfigurable drone swarm and what are the different hardware and software that make reconfigurability and elasticity possible.